%% file: main.tex
\documentclass[letterpaper, 10 pt, conference]{ieeeconf}  

\IEEEoverridecommandlockouts                              
\usepackage{xcolor}
\usepackage{comment}
\usepackage{graphics} 
\usepackage{epsfig} 
\usepackage{amsmath} 
\usepackage{amssymb}  
\usepackage{mathtools}
\usepackage{marginnote}
\usepackage{hyperref}

\usepackage{cite}
\usepackage{subcaption}
\usepackage{afterpage}
\usepackage{mathrsfs}

\newtheorem{definition}{Definition}

\title{\LARGE \bf
Reference-Steering via Data-Driven Predictive Control for Hyper-Accurate Robotic Flying-Hopping Locomotion} %

\author{Yicheng Zeng\mbox{*}, Yuhao Huang\mbox{*}, and Xiaobin Xiong 
\thanks{{\mbox{*} The authors contribute equally to this work.} The authors are with the Wisconsin Expeditious Legged Locomotion (WELL-Lab) at the University of Wisconsin-Madison. {Corresponding to  \tt\small xiaobin.xiong@wisc.edu}.
}%
\thanks{The experiment video can be seen here: \href{https://youtu.be/YM8VbjKEK9A}{https://youtu.be/YM8VbjKEK9A}}
}

\newtheorem{theorem}{Theorem}[section]

\begin{document}

\newcommand{\Kang}[1]{{\color{blue} #1}}

\newcommand{\acknowlege}[1]{\noindent{\textbf{Acknowledgement}: #1}}
\newcommand{\remark}[1]{\noindent{\textbf{#1}:}}

\newcommand{\ulparagraph}[1]{\noindent{\underline{#1}:}}

\newcommand{\block}[1]{\noindent{\textbf{#1}:}}
\newcommand{\TODO}[1]{{\color{blue} \textbf{TODO: #1}}}

\maketitle
\thispagestyle{empty}
\pagestyle{empty}

\begin{abstract}
State-of-the-art model-based control designs have been shown to be successful in realizing dynamic locomotion behaviors for robotic systems. The precision of the realized behaviors in terms of locomotion performance via fly, hopping, or walking has not yet been well investigated, despite the fact that the difference between the robot model and physical hardware is doomed to produce inaccurate trajectory tracking. To address this inaccuracy, we propose a referencing-steering method to bridge the \emph{model-to-real gap} by establishing a data-driven input-output (DD-IO) model on top of the existing model-based design. The DD-IO model takes the \emph{reference tracking trajectories as the input} and \emph{the realized tracking trajectory as the output}. By utilizing data-driven predictive control, we steer the reference input trajectories online so that the realized output ones match the actual desired ones. We demonstrate our method on the robot PogoX to realize hyper-accurate hopping and flying behaviors in both simulation and hardware. This data-driven reference-steering approach is straightforward to apply to general robotic systems for performance improvement via hyper-accurate trajectory tracking. 
\end{abstract}

\input{sections/intro}

\input{sections/preliminary}

\input{sections/method}

\input{sections/experiment}
\section{Conclusion and Future Work}
In conclusion, we propose a reference-steering approach via Data-Driven Predictive Control to improve trajectory tracking on robotic locomotion. We evaluated this approach both in simulation and on the hardware of the robot PogoX to realize both flying and periodic hopping locomotion. The results show significant tracking improvement for both locomotion behaviors, indicating a promising data-driven control approach that is augmented to the original model-based control design for general robotic systems. 

Our future work will be on extensions to nonlinear closed-loop dynamics and online updating of the Hankel matrices to build data-driven models that can adapt to system dynamics going through changes, e.g. locomotion over unknown and varying terrains. Additionally, we are interested in designing data-driven modeling and predictive control with augmentation to the model-based control and state estimation for general robotics tasks.  


\newpage
\bibliographystyle{IEEEtran}
\bibliography{reference}

\addtolength{\textheight}{-0cm}   

\end{document}

%% file: sections/intro.tex
\section{Introduction}

Model-based control design has been the core approach to realizing trustworthy dynamic, efficient, and safe behaviors on modern robotic systems, especially on flying~\cite{mahony2012multirotor,maaruf2022survey} and legged robots~\cite{wensing2023optimization}. The robot model being used, despite being either simplified or comprehensive such as the full-order Lagrangian dynamics model, describes the essential dynamics of how the input force/torques influence the states of the robot. Model-based controllers such as state feedback controllers~\cite{raibert1986legged, xiong20223, landry2016aggressive, pogox} and Model Predictive Control~\cite{bledt_mit_2018,ding2022orientation, li_versatile_2022} utilize the robot model to optimally plan and stabilize the trajectories of the robot for realizing locomotion tasks. 

The models derived from theoretical physics laws are expected to capture the actual dynamics of the robot to some good extent but never exactly, since the actual robot can have complex components and processes that may not be feasible to model, producing this \emph{model-to-real} gap. For instance, the robot linkages can have inaccuracy physical properties and they deform under load, the lower-level motor controls may not be fast enough to realize the desired torque or motor speed, and the sensing and computation loops have delays~\cite{siciliano2008springer, lynch2017modern}. Additionally, in real-world deployment, there will also be uncertainties on the hardware from the environment that we cannot model. Therefore, it is expected that in the model-based control design process, the realized trajectories can be different than the reference trajectories, yet within a reasonable range of tracking errors. 
 
\begin{figure}[t]
    \centering
    \includegraphics[width=\linewidth]{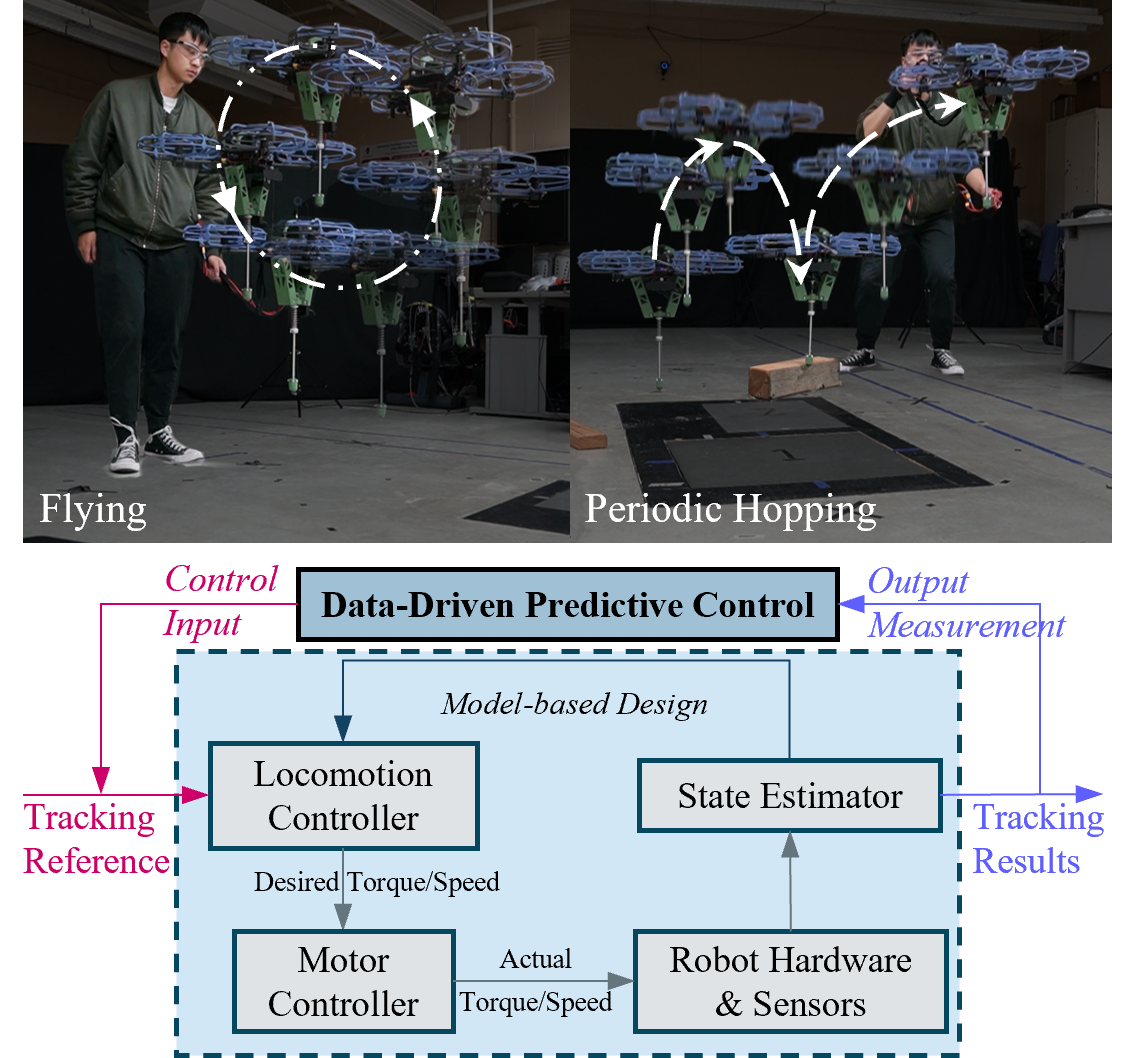}
    \caption{Overview of the data-driven reference-steering approach (bottom) and the application on PogoX (top).}
    \label{fig:overview}
    \vspace{-19.5pt}
\end{figure}

Data-driven and learning-based approaches~\cite{shi2019neural, markovsky2008data, dai2023data, li2024reinforcement, bianchini2023simultaneous, ferede2024end, coulson2019data, su2024leveraging, bellegarda2024robust, chen2024reinforcement, chang2020learning} have been the alternatives to address the problems of lack accurate models. Roughly speaking, they are concerned with data-driven models of the systems~\cite{shi2019neural, dai2023data, salzmann2023real, bianchini2023simultaneous}, controllers~\cite{bellegarda2024robust, li2024reinforcement}, or combinations of them~\cite{yang2020data, ferede2024end}. By and large, the data-driven and learning-based approaches either learn the controller or require re-synthesizing the control design based on the learned model, both of which abandon the original model-based control design that has proven theoretical properties. 

In this work, we aim to utilize the data-driven predict control (DDPC) techniques but keep the original model-based control design intact to address tracking errors caused by the model-to-real gap in robotic locomotion control. Fig.~\ref{fig:overview} illustrates our framework. We assume a model-based control design, presented by the dashed box, has been synthesized to realize robotic locomotion by performing trajectory tracking. The model-to-real gap causes errors between the reference and resultant trajectories. We then take a perspective of treating the reference trajectory as \emph{the input} and the resultant trajectory as {the output} of the dashed-boxed system, for which we apply the data-driven predictive control to address the model discrepancy to realize hyper-accurate trajectory tracking. In other words, we \emph{predictively steer the reference trajectory} (the input) so that the actual realized one (the output) follows the original target trajectory using a data-driven input-output (DD-IO) model. 

We leverage reference-steering via DDPC online to realize precise flying and periodic hopping on the robot PogoX~\cite{pogox,Kang2024fast} in both simulation and hardware. The original flying and hopping behaviors are realized by model-based controllers, which produce inaccurate tracking due to the model-to-real gap. We show that by utilizing reference-steering, DDPC bridges the gap and significantly improves the tracking accuracy. Moreover, to enable predictive control on hopping with hybrid dynamics, we present an artificial IO data generation process so that a uniform DDPC controller can be used to realize control over hybrid dynamics. The results validate our data-driven reference-steering and indicate a promising system-level data-augmented control design paradigm for complex robotic systems.

\section{Related Work}
The literature on data-driven or learning-based methods is very rich. Here, we mainly present the related work on developing data-driven predictive control (DDPC) and its application in robotics. DDPC is a recently popular development on data-driven control in the control community. It utilizes the behavioral system theory \cite{behaviroal_system_theory} to represent dynamical systems via data and then apply the Model Predictive Control (MPC) perspective for control. DeePC is an alternative acronym first used in~\cite{coulson2019data} with its widely accepted formulation. The recent development has been focused on extensions to high-dimensional systems~\cite{PEM-MPC_origin}, nonlinear dynamical systems~\cite{nonlinear_deepc}, and online data updates~\cite{online_deepc}. 

DDPC has been applied to robotics systems in the literature with a primary goal of using data to accurately capture unknown robot dynamics. For instance, \cite{deepc_quadcopter} utilized DDPC to predict accurate quadcopter dynamics. \cite{8768048} developed a data-driven error model to improve the performance of MPC on manipulators. \cite{PEM-MPC_quadruped} applied DDPC to better fit a single-rigid-body dynamics model for quadrupedal locomotion. \cite{li2024datadd} used DDPC to develop a template model for an exoskeleton that can easily learn and adapt to disturbances. In addition to modeling robot dynamics, DDPC can also be used to model disturbance forces and constraints that lack explicit models. \cite{10160914} has proposed a DDPC-based collaborative control method for modeling constraints of multi-agent robotic systems and the modeled dynamics are incorporated in a distributed trajectory planner. 



%% file: sections/preliminary.tex
\section{Preliminaries}
We begin by briefly introducing the behavioral system theory and the data-driven predictive control (DDPC) framework~\cite{coulson2019data, baros2022online,9109670, PEM-MPC_quadruped, li2024datadd}, which are the main control-theoretic tool utilized in this paper. We also succinctly review our previous model-based control of the robot PogoX~\cite{pogox}. 

\subsection{Behavioral System Theory}
The fundamental principles of behavioral systems theory \cite{behaviroal_system_theory} offer a mathematically rigorous method to represent an unknown Linear Time-invariant (LTI) system purely through its measured input-output data. The standard representation of a discrete-time LTI system where the state is denoted as $x_k \in \mathbb{R}^n$, the input as $u_k\in \mathbb{R}^m$ and the output as $y_k \in \mathbb{R}^p$ for $k\in \mathbb{Z}_{\geq 0}$ is given by: 
\begin{equation} \label{eq:linearDyn}
\begin{aligned}
x_{k+1} =  Ax_k + Bu_k,~
y_k = Cx_k + Du_k,
\end{aligned}
\end{equation}
where $A \in \mathbb{R}^{n\times n}, B\in \mathbb{R}^{n\times m}, C\in \mathbb{R}^{p\times n} \text{ and } D\in \mathbb{R}^{p \times m}$ are the state-space matrices. 
Given an input trajectory that starts at time $t_0$ and ends at time $t_1$, the sequence $u_{[t_0,t_1]} := [u_{t_0}^T, u_{t_0 +1}^T, \ldots, u_{t_1}^T]^T \in \mathbb{R}^{mT}$ has a length $mT$. Its Hankel matrix with depth $L$ can be defined as:
\begin{equation}
    \mathcal{H}_L(u_{[t_0,t_1]}) := 
    \begin{bmatrix}
    u_{t_0} & u_{t_0+1} &\ldots & u_{t_1 - L + 1\phantom{-}} \\ u_{t_0+1} & u_{t_0+2} &\ldots & u_{t_1 - L + 2} \\
    \vdots & \vdots & \ddots & \vdots \\
    u_{t_0+L-1\hspace{-1pt}} & u_{t_0+L} & \ldots & u_{t_1}\end{bmatrix} 
\end{equation}
\begin{definition}\label{PE}
    The input trajectory $u_{[t_0,t_1]}$ is said to be \textit{persistently exciting} (PE) of order $L$ if its corresponding Hankel matrix $\mathcal{H}_L(u_{[t_0,t_1]})$ has full row rank. 
\end{definition}

\begin{definition}
    An input-output pair $w(t) = (u(t),y(t))$ where $t\in\mathbb{Z}_{\geq 0} \text{ and } w(t)\in \mathbb{R}^{m+p}$ is said to be the \textit{input-output trajectory} of a discrete-time LTI system if for all $t\in \mathbb{Z}_{\geq 0}, \exists x: \mathbb{Z}_{\geq 0} \to \mathbb{R}^n$ such that \ref{eq:linearDyn} holds. 
\end{definition}

\begin{theorem}[Williem's Fundamental Lemma] \label{FundamentalLemma}
        For a~discrete-time LTI system described in~\eqref{eq:linearDyn}, assume $(A,B)$ is controllable. Let $(u_{[0,T-1]},y_{[0,T-1]})$ be a sequence of input-output trajectory with $T, L \in \mathbb{Z}_{\geq 0}$. If $u_{[0,T-1]}$ is persistently exciting of order $L+n$, then for a sequence of new input-output trajectory $(\bar{u}_{[0,L-1]},\bar{y}_{[0,L-1]})$ there always exists $g \in \mathbb{R}^{(m+p)\cdot(T - L + 1)}$ such that: 
    \begin{equation}
        \begin{bmatrix}\mathcal{H}_L(u_{[0,T-1]})\\ \mathcal{H}_L(y_{[0,T-1]}) \end{bmatrix} g =\begin{bmatrix}\bar{u}_{[0,L-1]} \\\bar{y}_{[0,L-1]} \end{bmatrix}.
    \end{equation}
\end{theorem} 
\vspace{8pt}


\subsection{Data-Driven Predictive Control} \label{deepc_intro}
This section briefly overviews the formulation of data-driven predictive control. As for Williem's Fundamental Lemma, we can assume that $L = T_{ini} + T_f$, where $T_{ini}$ denotes the length of the estimation horizon, and $T_f$ denotes the prediction horizon. The estimation horizon should be larger than the lag of the system to uniquely determine the initial condition of the unknown LTI system for a given trajectory sequence $(\bar{u}_{[0,L-1]},\bar{y}_{[0,L-1]})$. 
\begin{definition}
The \textit{lag} $\ell$ of a discrete-time LTI system is defined as the smallest integer $k$ such that the observability matrix $\mathcal{O}_k$ has rank $n$. 
\end{definition}

\begin{theorem}
Let $T_\text{ini} \geq \ell$, and let $[u_\text{ini}^T, u^T, y_\text{ini}^T, y^T]^T$ where $(u_\text{ini},y_\text{ini})\in \mathbb{R}^{T_\text{ini}\cdot(m+p)}, (u,y)\in\mathbb{R}^{T_f\cdot(m+p)}$ be a sequence of input-output trajectories. Then there exists a unique state vector $x_\text{ini} \in \mathbb{R}^n$ such that: $y = \mathcal{O}_{T_f} x_\text{ini} + \mathcal{T}_{T_f} u$, where $\mathcal{T}_{T_f}$ is the Toeplitz matrix~\cite{coulson2019data} of the LTI system. 
\end{theorem}

The previous definitions and theorems give rise to the division of Hankel matrices of input-output trajectories into past data for uniquely determining initial conditions and prediction part in future horizons for control: 
\begin{equation}
       \mathcal{H}_L(u_{[0,T-1]}) = \begin{bmatrix}
           U_p \\ U_f
       \end{bmatrix}, \mathcal{H}_L(y_{[0,T-1]}) = \begin{bmatrix}
           Y_p \\ Y_f
       \end{bmatrix}, 
\end{equation}
where $U_p \in \mathbb{R}^{mT_\text{ini}\times(T-L+1)}, U_f\in \mathbb{R}^{mT_f\times(T-L+1)}, Y_p \in \mathbb{R}^{pT_\text{ini}\times(T-L+1)} \text{ and } Y_f \in \mathbb{R}^{pT_{f}\times(T-L+1)}$, and the subscripts $_p$ and $_f$ denote the data in the past and future, respectively. Then a Data-Driven Predictive Control (DDPC) problem can be formulated as a quadratic program (QP) as in \cite{coulson2019data}: 
\begin{align}
 \min_{(g,u,y,\sigma_y)} &  \|y-y^{\mathrm{des}}\|_Q^2+\|u\|_R^2 
 +\lambda_g\|g\|^2+\lambda_\sigma\|\sigma_y\|^2,  \label{eq:DDPC}\\
\mathrm{s.t.}   &   \begin{bmatrix}U_p\\Y_p\\U_f\\Y_f\end{bmatrix}g=
 \begin{bmatrix}u_\text{ini}\\y_\text{ini}-\sigma_y\\u\\y\end{bmatrix} 
 ,~u\in\mathcal{U}, \quad y\in\mathcal{Y}, \nonumber
 \end{align} 
 where $Q,R\succ 0$ are weighting matrices for inputs $u\in \mathbb{R}^{m T_f}$ and outputs $y\in \mathbb{R}^{p T_f}$, and $\lambda_g$ and $\lambda_\sigma$ act as penalty terms for $g$ and $\sigma_y$. Note that $\sigma_y$ is commonly introduced to represent measurement noise on the output. $\mathcal{U},\mathcal{Y}$ denote the feasible sets of the input and output trajectories. 

\subsection{Flying-Hopping Control of PogoX} \label{pogox_control_pre}
Here we briefly review our previous control design~\cite{pogox} for realizing flying-hopping on PogoX. PogoX is a hybrid robot that combines a flying quadrotor with a pogo-stick, which enables terrestrial hopping behaviors. The current version of the robot~\cite{Kang2024fast} weighs 3.65kg with a height of 0.53m. It is fully equipped with various sensors and computation power to perform hybrid locomotion in a real-world environment. 

\ulparagraph{Flying Control} Since the core body of the robot is a quadrotor, flying has been directly realized by using its off-the-shelf on-board PX4 flight controller~\cite{PX4}, which has cascaded linear control loops to sequentially control angular rate, body orientation, and then linear velocities and linear positions. 

\ulparagraph{Hopping Control} The previous hopping controller is composed of two main components: a vertical energy controller that is responsible for maintaining a certain apex height, and a horizontal velocity controller that aims for velocity tracking. The vertical energy controller is designed as a control Lyapunov Function based Quadratic Program (CLF-QP controller)~\cite{pogox,ames2014rapidly} which stabilizes the vertical mechanical energy to a desired value.  
The horizontal velocity controller utilizes a step-to-step (S2S) dynamics~\cite{pogox,xiong20223} based stepping control where the nominal hopping gait is optimized from the vertical energy-controlled spring-loaded inverted pendulum (SLIP) model.
Readers can refer to~\cite{pogox} for details. 

\remark{Remark}\label{remark} The hopping controller uses the Euler angles and total thrust forces rather than the motor speeds as the control input. This is primarily because the on-board flight controller provides an ``off-board" mode in which the users can command the Euler angles and total thrust force, which are then realized by its internal PI/PID controllers, and the users can only adjust the control gains within certain ranges. This layered controller architecture is similar to other robotic systems where the users can command desired torque or joint velocity, which are realized by the proprietary current and joint controllers from the vendor. In practice, the proprietary controller can have degraded tracking performance if they are not tuned for the target application; in this case, the users cannot modify the proprietary controller part to improve tracking. This also happened to PogoX since the flight controller was designed for quadrotor-flying, thus the proprietary controller oftentimes provides unsatisfying tracking performance. This is one of the motivations of our research here: we aim to steer the reference rather than re-design the existing control architecture.   



%% file: sections/method.tex
\section{Reference-Steering for Flying}
In this section and the next, we describe the reference-steering via DDPC for realizing flying that has continuous dynamics and hopping that has hybrid dynamics. For PogoX which has a ``flight controller" of quadrotors, the choice of the tracking trajectories can either be the Cartesian position with yaw angle or the vertical thrust force with orientation, which are in different operation modes (on-board v.s. off-board mode). Switching the modes during locomotion is prohibited due to the hardware. With an aim to use one mode to enable versatile locomotion, we opt to use the ``off-board" mode. The reason will be highlighted in the next section.

\subsection{Flying Dynamics and Controller}
\ulparagraph{Robot Dynamics}  
The flying dynamics of PogoX is governed by~\cite{Quad_3D}: $m\ddot{\mathbf{r}} =\begin{bmatrix}0, 0, -mg\end{bmatrix}^T+ \mathbf{R} \begin{bmatrix}0, 0, u_1\end{bmatrix}^T, \mathcal{I}\dot{\omega}  =\begin{bmatrix}u_2,u_3,u_4\end{bmatrix}^T-\omega\times \mathcal{I}\omega$,
 where $u_1$ is the total thrust force, $u_2,u_3,u_4$ represent the moments generated on the body frame, $\omega$ represents the angular velocity in body frame, and $\{\mathbf{r}, \mathbf{R}\}$ represent the position and orientation of the robot in the world frame, respectively. The thrust force and moment generated by the rotor can be modeled as $\mathcal{F} = k_F \omega_\text{rotor}^2$, where $k_F$ is constant determined by the propeller, and $\omega_\text{rotor}$ is the rotor velocity. The motor dynamics and its controller are all linear. 
 

\ulparagraph{Flying Control Design} 
When designing flying control, $(\mathbf{r},\psi)$ is commonly used as the output of trajectory planning and control, where $\psi$ is the yaw angle of the robot. With an eye to use consistent IO for both flying and hopping, here we choose the Euler angles and total thrust, $(\psi,\theta,\phi,F^{\sum}_{\text{thrust}})$, as the output of the "flight controller" in its ``off-board" mode. To realize flying in the Cartesian space, we first design a linear PD controller to stabilize the vertical position. The horizontal positions are stabilized by directly calculating the corresponding desired Euler angles. We assume the robot does not turn for simplicity, i.e. $\psi = 0$. Since the roll and pitch angles directly relate to the leg angle of the robot, we refer to the model-based controller design as ``Height and Leg Angle Controller", as illustrated in Fig.~\ref{fig:control_layers_pogoX}, which realizes both flying and hopping given corresponding desired trajectories.

\subsection{Reference-Steering via DDPC}
We first show that the closed-loop input-output (IO) dynamics of PogoX during flying can be approximated by an LTI system, which rationalizes the application of DDPC for reference-steering. We then present a computationally efficient DDPC for realizing online reference-steering. 

\begin{figure}
    \centering
    \includegraphics[width=0.9\linewidth]{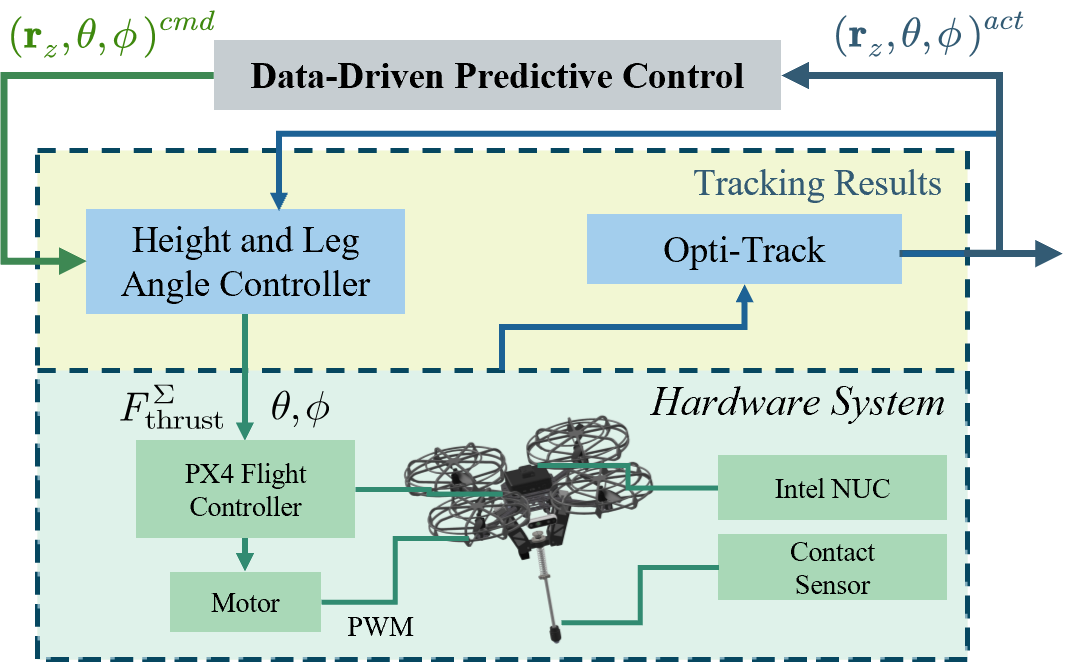}
    \caption{The control architecture of PogoX.}
    \label{fig:control_layers_pogoX}
    \vspace{-10pt}
\end{figure}

\ulparagraph{Closed-loop IO Dynamics} During the flying and hopping operation of the robot, it is reasonable to assume: (a) the robot operates with small roll and pitch angles, where $\cos \phi \approx 1, \sin \phi \approx \phi, \cos \theta \approx 1, \sin \theta \approx \theta$; and (b) the cross product term of the angular velocity is relatively small, which yields a linear robot dynamics. The closed-loop IO dynamics of the dashed-box diagram in Fig.~\ref{fig:control_layers_pogoX} is thus simplified to a linear system, as the motor dynamics, and controllers are all linear. Additionally, many unmodeled dynamics and processes on the robot are approximately linear as well, such as the blade flapping, deformation of the limbs, and internal electronic processes. Therefore, it is reasonable to use the IO of the closed-loop system in DDPC for reference-steering. 

\remark{Efficient DDPC Formulation} To apply the DDPC, we first off-line generate sufficient IO data of the closed-loop system to realize flying behaviors to construct the Hankel matrices offline. Then, we use the Hankel matrices to formulate \eqref{eq:DDPC} for reference-steering. The offline data may result in large Hankel matrices, which thus prevent real-time computation. Additionally, the sensor noises can lead to infeasibility. To address these problems, we introduce an extension to the PEM-MPC formulation of DDPC \cite{PEM-MPC_origin} to improve the computational efficiency and robustness to sensor noise. 

We consider to first solve for the linear predictor $g$ and the slack variable $\sigma_y$ of this optimization problem: 
\begin{align}
 \min_{g, \sigma_y} \quad ||g||^2 + ||\sigma_y||^2, \quad \text{s.t.} \begin{bmatrix} U_p \\ Y_p \\ U_f \end{bmatrix} g = \begin{bmatrix} u_{\text{ini}} \\ y_{\text{ini}} + \sigma_y \\ u \end{bmatrix},
\end{align}
the solution of which is ($^{\dagger}$ represents the pseudo-inverse):
\begin{equation}
    \begin{bmatrix}
        g \\ \sigma_y
    \end{bmatrix} = \begin{bmatrix} U_p&0 \\ Y_p& I \\ U_f&0 \end{bmatrix}^{\dagger}\begin{bmatrix} u_\text{ini} \\ y_\text{ini} \\ u \end{bmatrix}.~\text{Let}~\mathcal{G} =\begin{bmatrix} Y_f & 0 \\ 0 & I\\ \end{bmatrix} 
\begin{bmatrix} U_p&0 \\ Y_p& I \\ U_f&0 \end{bmatrix}^{\dagger}.\nonumber
\end{equation}
The reference-steering problem can thus be formulated as: 
\begin{align}
\min_{u,y,\sigma_y}& ||y-y^\text{des}||^2_Q + ||u||_R^2 + \lambda_\sigma ||\sigma_y||^2,   \label{eq:ddpc_flying} \tag{DDPC-Flying}
\\ \text{s.t.} \quad & \begin{bmatrix} y \\ \sigma_y \end{bmatrix} = \mathcal{G}\begin{bmatrix}u_\text{ini} \\ y_\text{ini} \\ u \end{bmatrix}, u\in\mathcal{U}, y\in\mathcal{Y},
 \nonumber       
\end{align}
 where $y^\text{des}$ is the desired trajectory, $u$ and $y$ are commanded trajectory and realized trajectory. $\mathcal{G}$ is computed offline, and solving this QP online yields the control input $u$ to steer the realized output $y$ towards the desired one $y^\text{des}$. 



\section{Reference-Steering for Periodic Hopping}
We now present our approach of using reference-steering via DDPC to enable periodic hopping locomotion on PogoX. Its hybrid dynamics presents a severe challenge from both a theoretical and a practical perspective on using the continuous predictive control approach for trajectory stabilization via either MPC or DDPC. We thus innovate an "artificial IO" trajectory generation process on the ground phase to enable continuous predictive control via DDPC in the aerial phase.  

\subsection{Hybrid Dynamics and Hopping Control}
\ulparagraph{Hybrid Dynamics of Hopping} The dynamics of hopping is hybrid, as it has two different phases: an aerial phase and a ground phase. The dynamics in the aerial phase are identical to the quadrotor dynamics. On the ground, the robot is subjected to ground reaction force due to contact, and we assume the foot does not slip during contact, which yields a holonomic constraint. The dynamics in the ground phase can thus be described by a differential-algebraic equation. The transition from the aerial phase to the ground phase is governed by a discrete impact map, while the transition from the ground phase to the aerial phase occurs smoothly. The complete hybrid dynamics description can be seen in~\cite{pogox}. 

\ulparagraph{Hopping Control Design} To enable dynamic hopping behaviors with consistent IO of flying, we modify our previous hopping control in~\cite{pogox}. Instead of stabilizing the energy to the desired one in the aerial phase, we directly track the vertical height of the robot. The leg angles, equivalent to the roll and pitch angles of the robot body, are used for control which is the same as the previous one in~\cite{pogox}. The desired vertical trajectory is optimized using the robot dynamics for vertical hopping. The leg angles can perturb and stabilize the vertical hopping into hopping with different horizontal velocity via step-to-step (S2S) dynamics based control~\cite{xiong20223}. Since the ground phase is very short and abrupt, we simply let the desired output values be the current measured ones, which can preserve the natural dynamics of the spring leg. 

\remark{Choice of IO} The alternative choice of IO can be the vertical height plus the horizontal position of the robot. It may be able to produce dynamic hopping if the horizontal desired trajectory is well-designed or optimized, which thus requires additional trajectory optimization to realize hopping with different horizontal velocities. While for hopping with leg angle as outputs, variable horizontal velocities can be produced directly~\cite{raibert1986legged, xiong20223} without additional trajectory optimization. Therefore, in our application, we select the vertical height and body orientation as outputs. This is the reason we use these outputs for flying rather than the Cartesian positions.



\subsection{Reference-Steering via DDPC}
\begin{figure}[t]
    \centering
    \includegraphics[width=0.99\linewidth]{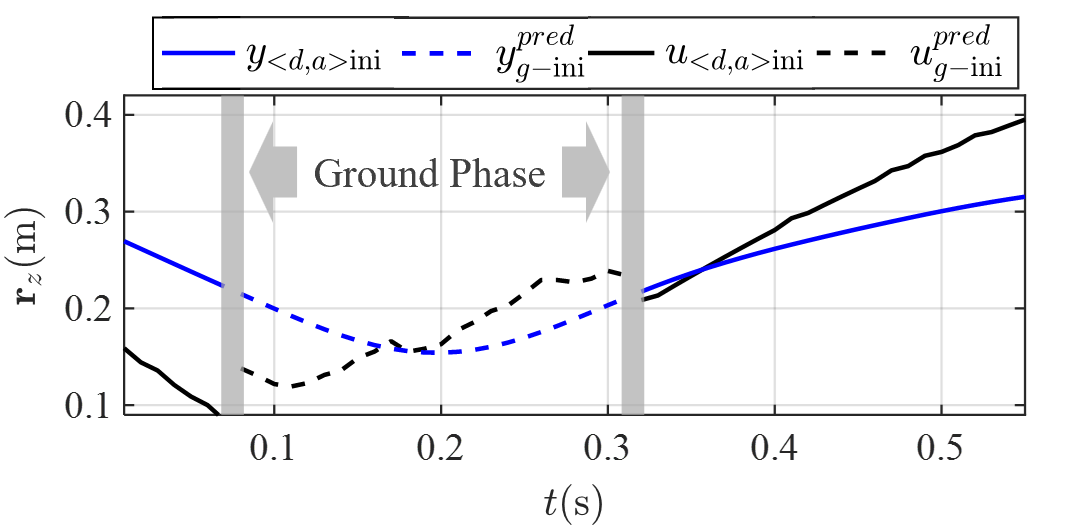}
    \caption{Illustration of artificial ground phase trajectory generation, where $u_{<d,a>\text{ini}}$ and $y_{<d,a>\text{ini}}$ represent the collected input and output data, and $u_{g-\text{ini}}^{pred},y_{g-\text{ini}}^{pred}$ represent the generated IO data of the ground phase.
    }
    \label{fig:artificial_IO}
    \vspace{-6pt}
\end{figure}


\ulparagraph{Artificial IO Generation} This model-based hopping control design can produce dynamic hopping. Yet, due to the model discrepancy and the tracking errors on the desired outputs, the realized hopping can have errors in its hopping height and velocity. We then apply the reference-steering with DDPC to improve the trajectory tracking results. Since the hopping dynamics is hybrid, the Hankel matrix built for flying tasks cannot be directly applied to hopping, and the IO data of hopping cannot be used for hopping since they are from two different dynamics. Fortunately, the IO data of the aerial phase still represents the dynamics of the aerial phase, which is the phase we apply control. Therefore, we focus on generating artificial IO data in the ground phase that represents the aerial dynamics. With artificial IO data in the ground phase and the real IO data in the aerial phase, we have continuous IO trajectories of the hopping behavior, where the IO data only represents the aerial dynamics. We can then apply DDPC in the aerial phase that has the approximately linear IO dynamics shown in the previous section.  

Then our goal is to design $u_{g-\text{ini}}, y_{g-\text{ini}}$ that artificially represent the ground trajectories using aerial dynamics. For each hopping, we formulate this QP problem: 
 \begin{align}
&\min_{u_{g-\text{ini}}, y_{g-\text{ini}}, g} ||g||^2,  \\
\text{s.t.}& \begin{bmatrix}U_{p}\\Y_{p}\end{bmatrix}g=\begin{bmatrix}
u_{<d,g,a>\text{ini}}\\ y_{<d,g,a>\text{ini}}\\
\end{bmatrix}, u\in\mathcal{U}, y\in\mathcal{Y}, \nonumber
\end{align}
where $U_p, Y_p$ are the Hankel matrices representing the aerial dynamics, and the subscripts $d, g, a$ denote the descending, ground, and ascending phases, respectively. Fig.~\ref{fig:artificial_IO} illustrates that by selecting an appropriate length for the flying initial IO trajectories $T_{d-\text{ini}}$ and $T_{a-\text{ini}}$, the artificial ground IO trajectories $u_{g-\text{ini}}, y_{g-\text{ini}}$ can be uniquely determined. If we collect $n$ trajectories from $n$ hopping cycles, a new Hankel matrix, which incorporates artificial ground phase initial trajectories, can be expressed as $H_G = [H_{G1} | H_{G2} | \ldots | H_{Gn}]$ where $H_{Gk}$ is the Hankel matrix of k-th trajectory $[u^k_{d-\text{ini}}, u^k_{g-\text{ini}}, u^k_{a-\text{ini}}, y^k_{d-\text{ini}}, y^k_{g-\text{ini}}, y^k_{a-\text{ini}}]$ with $k\in \mathbb{Z}_{[1,n]}$.

\remark{DDPC for Periodic Hopping} With a relatively short and fixed ground phase, we can infer that the length of the initial trajectories exceeds that of the ground phase. We thus implicitly assume the length of the initial trajectories before and after the ground phase is longer than the system lag \ref{deepc_intro} such that system states at lift-off and touch-down are uniquely determined. The DDPC problem, which includes the ground phase, can then be described as follows:
 \begin{align}
\min_{u,y,\sigma_y}& ||y-y^\text{des}||_Q^2 + ||u||_R^2 + \lambda_{\sigma}||\sigma_y||^2, \tag{DDPC-Hopping}\\
\text{s.t.}& \begin{bmatrix}y\\\sigma_y\end{bmatrix}=\mathcal{G}\begin{bmatrix}u_{<d,g,a>\text{ini}}\\ y_{<d,g,a>\text{ini}}\\u\end{bmatrix}, u\in\mathcal{U}, y\in\mathcal{Y}. \nonumber 
\end{align}
This formulation is then uniform with that in \eqref{eq:ddpc_flying}. Similarly, $\mathcal{G}$ is computed offline, and then this QP problem can be computed online in real time. 




%% file: sections/experiment.tex
\section{Evaluation}
In this section, we will present the results of our approach for realizing hyper-accurate trajectory tracking on the robot PogoX in both simulation and hardware. The hardware control is realized in ROS environment. For indoor testing, the actual trajectories are measured by an Opti-track system that is composed of 12 Prime-13 cameras. The proposed DDPC problems are implemented in C++, and the corresponding QPs are solved using OSQP~\cite{stellato2020osqp} with warm-starting at 200 Hz, which matches the computation frequency of the reference trajectory planners and the low-level controllers. The simulation is implemented using MATLAB ODE function with event-triggering, where the control part is set up similarly to match that of the hardware.


To mitigate the potential of non-persistent excitation in the input signal, which is required by~\ref{FundamentalLemma}, we adopt the approach in \cite{deepc_quadcopter} by applying a small scaled pseudo-random binary sequence (PRBS) noise to the input signals while minimizing its influence on the control quality. 

The previous version of PogoX~\cite{Kang2024fast} has a thrust-to-weight ratio (TWR) estimated at $\sim 0.82$. To enable flying in the work, we take the batteries off the robot to have a TWR slightly over 1. All the flying and hopping data for the Hankel matrix construction were collected under this condition. 



\begin{figure}[t]
    \centering
    \includegraphics[width=1\linewidth]{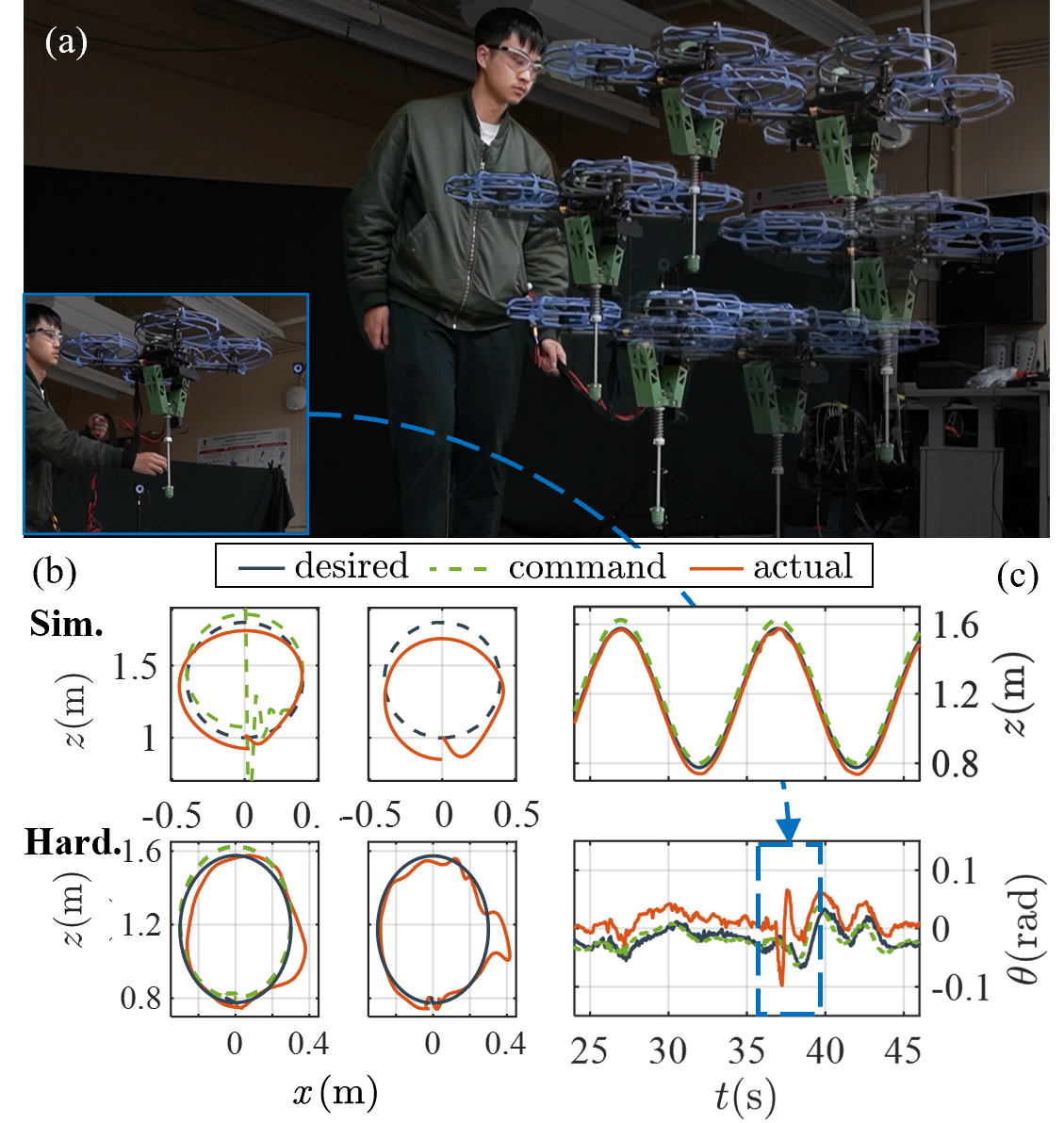}
    \caption{Results on flying: (a) depicts PogoX in flying and disturbance injection; (b) compares the quality of trajectory tracking in $x-z$ plane; and (c) demonstrates how DDPC responds to disturbances in both height and leg angle tracking.}
    \label{fig:hardware_fly}
    \vspace{-2pt}
\end{figure}

\subsection{Reference-steering for Flying}
\begin{figure}[t]
    \centering
    \includegraphics[width=1\linewidth]{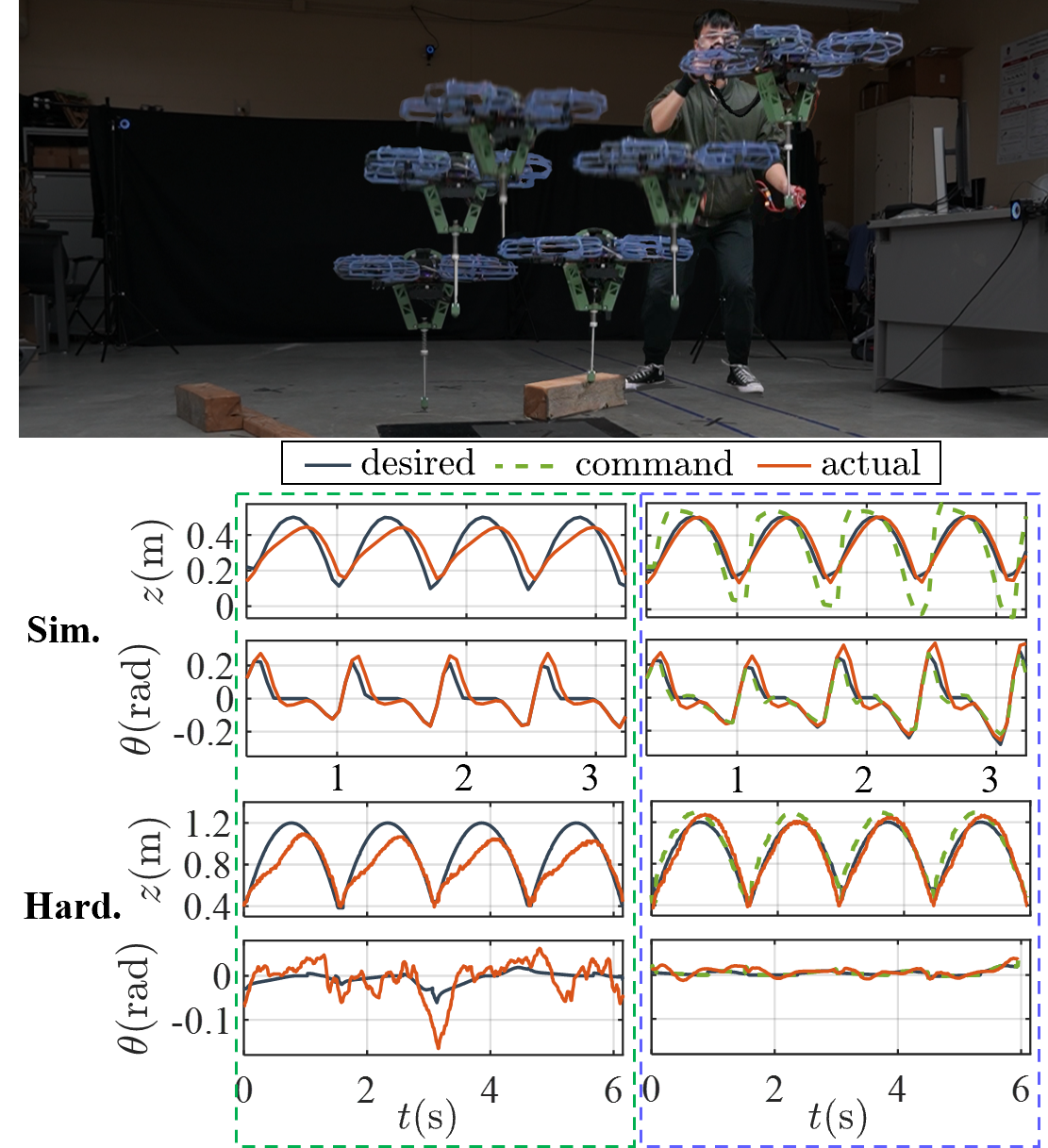}
    \caption{Results on periodic hopping: trajectories without reference-steering (green) and with reference-steering (blue).}
    \label{fig:hopping}
    \vspace{-4pt}
\end{figure}

We first present the results of applying reference-steering to the flying of PogoX. The input data used to construct the Hankel matrices is collected on the desired $z$ and $\theta$ trajectories that are mapped from RC control inputs. The output data comes from the direct measurement of the robot states. This data collection intends to capture the general IO dynamics in the air rather than those of specific tasks for following certain trajectories. The lower-level controller and its feedback gains during data collection and experiment remain the same. The robot is then commanded to follow an ellipsoid trajectory in its sagittal plane where the desired $x-$ direction position is transformed to the desired leg angle via a PD controller. Since the reference trajectories are designed to be relatively short, the initial trajectory length is set to 20 to sufficiently account for system lag, while the prediction horizon is kept relatively small to 15 to enhance computational efficiency.


\remark{Simulation} The height and leg angle reference tracking performance can be seen in Fig. \ref{fig:hardware_fly}. To challenge the controller, the robot mass in the controller is set to be 200g less than its real value. As shown in the $x-z$ plot, the quadrotor failed to follow the correct trajectory due to the modeling error, and reference-steering via DDPC steers it back to the desired height. Note that to enhance tracking in the $x$ direction, one can always add $x$ position in the output to formulate the dynamics with input as reference $\theta$ and $x$.

\remark{Hardware} The same controller and desired trajectory are used on the hardware with additional disturbance forces applied to the robot leg to disrupt both height and leg angle tracking. The comparison of the robustness of the controller, with and without DDPC, is shown in Fig \ref{fig:hardware_fly}. Despite we cannot quantitatively measure the applied disturbance for each experiment, DDPC is showing better performance for both regular tracking and disturbance rejection. 


\subsection{Reference-steering for Periodic Hopping}
We now present the results of the periodic hopping behaviors. During both the data collection and experimental evaluation, the robot is commanded to perform periodic hopping, aiming to realize a desired apex height and a target horizontal velocity. To ensure adequate coverage of both the descent and ascent phases during the aerial stage in predictive control, the prediction horizon is set to 25, and the initial trajectory length is maintained at 20. This allows the horizon to encompass a complete hopping cycle.


\remark{Simulation} Fig. \ref{fig:hopping} illustrates the hopping behavior performance in simulation, both with and without DDPC. It is evident that with DDPC, height tracking is significantly improved, while the leg angles maintain sufficiently accurate tracking performance. With DDPC, the reference trajectories are steered to higher targets to account for the tracking inaccuracies during lift-off.

\remark{Hardware} The height and leg angle reference tracking performance can be seen in Fig. \ref{fig:hopping}. Similar to the simulation results, the height and leg angle tracking have been significantly improved, leading to better overall stabilization of step-to-step (S2S) dynamics and hopping behaviors. The steered output demonstrates that DDPC can predict the mismatch between the reference and actual trajectories, adjusting the reference trajectory to better track the desired trajectories. 